\documentclass{article}

\usepackage{arxiv}

\usepackage[utf8]{inputenc} % allow utf-8 input
\usepackage[T1]{fontenc}    % use 8-bit T1 fonts
\usepackage{hyperref}       % hyperlinks
\usepackage{url}            % simple URL typesetting
\usepackage{booktabs}       % professional-quality tables
\usepackage{amsfonts}       % blackboard math symbols
\usepackage{nicefrac}       % compact symbols for 1/2, etc.
\usepackage{microtype}      % microtypography
\usepackage{lipsum}
\usepackage{graphicx}
\graphicspath{ {./images/} }

\title{Unsupervised Paraphrase Generation using Pre-trained Language Models}

\author{
   Chaitra Hegde \\
   Fidelity Investments \\
   Boston, MA \\
   \texttt{chaitra.vishwanathahegde@fmr.com} \\
   \And
   Shrikumar Patil \\
   Fidelity Investments \\
   Boston, MA \\
   \texttt{shrikumarrajendra.patil@fmr.com} \\
}

\begin{document}
\maketitle
\begin{abstract}
Large scale Pre-trained Language Models have proven to be very powerful approach in various Natural language tasks. OpenAI's GPT-2 \cite{radford2019language} is notable for its capability to generate fluent, well formulated, grammatically consistent text and for phrase completions. In this paper we leverage this generation capability of GPT-2 to generate paraphrases without any supervision from labelled data. We examine how the results compare with other supervised and unsupervised approaches and the effect of using paraphrases for data augmentation on downstream tasks such as classification. Our experiments show that paraphrases generated with our model are of good quality, are diverse and improves the downstream task performance when used for data augmentation. 
\end{abstract}

% keywords can be removed
%\keywords{First keyword \and Second keyword \and More}

\section{Introduction}
Paraphrasing is well known problem in NLP which has wide gamut of applications including data augmentation, data curation, intent mapping, semantic understanding. Paraphrasing can be used to generate synthetic data to augment existing scarse datasets for training, and to enhance performance of downstream tasks such as classification. It can also be used to extend existing datasets to more variations, so that it becomes more generalizeable, because of the linguistic variability added by paraphrasing. 

The supervised approach towards paraphrasing suffers from lack of domain specific parallel data. The unsupervised approaches towards paraphrasing is heavily reliant on machine translation which still requires bilingual parallel data corpus. Also, performance of machine-translation based back-translation depends on performance constituting machine translation systems. Both supervised approach and machine translation-based paraphrasing suffers when there is a severe domain shift during inference \cite{koehn-knowles-2017-six}. 

GPT-2 \cite{radford2019language} is exceptional at language generation. It predicts the next token in the sequence given all the tokens before it, i.e it optimizes for $P(X_i|X_<i)$. The pre-trained version of GPT-2 generates bland text without any goal. By making the model to generate Target T by conditioning on the Source S, the language generation capability of GPT-2 can be utilized for generating meaningful text. 

In this paper, we propose a novel unsupervised paraphrasing technique using GPT-2. As this is a unsupervised approach, the model can be trained on the domain specific independent sentences at hand and generate paraphrases without suffering from domain shift. We formulate the unsupervised paraphrasing task as sentence reconstruction task from corrupted input. From a sentence, we omit all the stop words to form a corrupted sentence, lets call it Source S, and the original sentence is used as Target T. We use GPT-2 to generate the Target sentence given Source. i.e $P(T|S)$

\section{Related Work}

The task of textual paraphrasing has been of high interest in NLP research and has been tackled using variety of approaches. Some earlier approaches towards paraphrasing like \cite{barzilay-lee-2003-learning}, \cite{quirk-etal-2004-monolingual} and \cite{duboue-chu-carroll-2006-answering} were mostly based on linguistic structures. In last few years, there has been a lot of work on framing generation of paraphrases as a Seq2seq task \cite{10.5555/3298023.3298027}\cite{prakash-etal-2016-neural}\cite{su-yan-2017-cross}.
Paraphrasing has also been explored as a means to improve Question answering in \cite{Buck2017AskTR} and \cite{Dong2017LearningTP} by making the training more robust by incorporating reformulations. 

Deep generative models and discrete latent structures for paraphrasing \cite{Fu2019ParaphraseGW} have also been explored, and \cite{Gupta2018ADG} proposed a combined the approaches of VAE and Seq2Seq model, by conditioning both encoder and decoder on the source sentence, Else ways, \cite{Li2018ParaphraseGW} and \cite{nomoto-2019-generating} involve reinforcement learning.\cite{qian-etal-2019-exploring} suggests using multiple generators to generate more diverse and robust variations, whereas \cite{Iyyer2018AdversarialEG} investigates generating paraphrases that conform to a certain target syntactic form.

Unsupervised paraphrasing is picking up a lot of pace in paraphrasing related research work. Approaches such as \cite{Bowman2016GeneratingSF} and \cite{Miao2018CGMHCS} employ variational autoencoders for sampling paraphrases from learned latent space. \cite{roy-grangier-2019-unsupervised} use Vector-Quantized AutoEncoders \cite{Oord2017NeuralDR} based model, claiming ability to generate more diverse while semantically closer sentences from input. \cite{Liu2019UnsupervisedPB} formulate paraphrasing as a stochastic searching problem and involve unsupervised searching algorithm, simulated annealing to generate paraphrases. Difference between these unsupervised approaches, and our approach is that, we aim to explore the power of large-scale pre-trained language model, which due to the sheer size and breadth of corpus they are trained on, perform extremely well on downstream tasks. As these models are primarily trained in a way to effectively learn to predict next word by understanding context, we explore it’s effectiveness in paraphrasing problem setup. The only other work to this end involving use of pre-trained large language model for paraphrasing is in \cite{Witteveen2019ParaphrasingWL}. Our work, although being similar to architecture proposed in \cite{Witteveen2019ParaphrasingWL}, the approach taken is completely different, as they fine-tune GPT-2 on labelled paraphrasing dataset, by contrast, we aim to do it completely unsupervised, by making the model learn more on reconstruction task rather than learn from labelled paraphrasing data. The evaluation in \cite{Witteveen2019ParaphrasingWL} is also significantly different, we run exhaustive quantitative evaluation and compare with reported standard metric numbers from other methods, including supervised and unsupervised methods, on same datasets, to better demonstrate its effectiveness, along with evaluation of paraphrasing quality, diversity and correctness.

\section{Methodology}
\subsection{Paraphrase Generation using GPT-2}

Paraphrasing is a technique of generating a Target sentence T for a the reference sentence P where the newly generated Target Sentence T is semantically similar to reference sentence P.

GPT-2 \cite{radford2019language} is a auto-regressive model which has achieved state-of-the art result in many of the benchmark tasks. It has up-to 1.5 Billion parameters and was trained simply to predict the next word in 40GB of Internet text. We utilize Huggingface Transformers Library \cite{Wolf2019HuggingFacesTS} to finetune the GPT-2 model on a sentence reconstruction task to generate paraphrases.

\subsection{Data Preparation}
We take unsupervised approach towards Paraphrase generation and hence no parallel corpora for paraphrasing has been utilized, instead we create a dummy dataset for sentence re-construction task. We corrupt the input sentence by removing all the stop words, further we randomly shuffle and replace the remaining words 20\% of the time. We call the corrupted sentence as source sequence S and the original uncorrupted sequence as target T. Goal is to reconstruct the sentence T from the keywords/corrupted sentence S, $P(T|S)$.  

$A: Set\ of\ stop\ words$ ; $T: Sentence$

$S: T-A$; 

$R: Paraphrase/Reconstructed\ Sentence$;

Our stop word set has total of 252 words. After removing the stop words from sentences to form Source S, we encourage model to generate diverse paraphrases by shuffling the words 20\% of the time. To encourage the model to have new words in the reconstructed sentence that were not in the original sentence, 20\% of the words are randomly replaced with a synonym using syn-net.

\subsection{Model Fine-tuning for Paraphrase Generation}
We utilize pre-trained GPT-2 checkpoint from Huggingface Library and we further finetune the model on Sentence Reconstruction task. We concatenate the Source and Target sequence separated by a special symbol to form the input sequence X. Let $X$=$x_0, x_1, x_2...x_m,[sep], x_{m+1},...x_n$. We denote the Source sequence as $S$=$x_0, x_1,...x_m$ and Target sequence as $T$=$x_{m+1}, x_{m+2},...x_n$. Special character $[SEP]$ is added to distinguish between source tokens and target tokens and it serves as a promt during inference.

\begin{table*}[t!]
\centering
  \caption{Paraphrasing Examples : Output from our model}
  \label{tab:freq}
  \begin{tabular}{cc}
    \hline \hline
    Type&Sentence\\
    \hline \hline
    In & how do you send a private message to someone you're following on quora?  \\
    Corrupted & send private message following quora ?   \\
    Out & how do i send a private message through quora?\\
    Out & how do i send a private message to a few of my friends on quora? \\
  \hline
  In & do you believe donald trump can make america great again?  \\
  Corrupted & believe donald trump america great ?   \\
  Out & do you believe donald trump is doing everything that he can to make america great again?   \\
    \hline
    In & if we see something in our dreams and it happens to come out true  after few days, what does that mean?  \\
    Corrupted & dreams happens come out true after days, mean?  \\
    Out & if dreams happens to come out true in a few days, what does it mean? \\
    \hline
    In & what do i gift my boyfriend for his birthday?  \\
    Corrupted & gift boyfriend birthday?  \\
    Out & what gift can i get my boyfriend for his birthday?  \\
    Out & what gifts should i give my boyfriend on his birthday? \\
    \hline\hline

\end{tabular}
\end{table*}

\subsection{Candidate Paraphrase Generation and Scoring}
To generate more than one paraphrase for a given input sentence, we use top-k sampling \cite{fan-etal-2018-hierarchical}. We generate 10 paraphrases for every input sentence, lets calls the paraphrased sentence as R. We eliminate the ones that are same as original sentence T or is different by only few characters. To make sure that the paraphrases are semantically similar to the input sentence T, we use Sentence Transformers Library \cite{reimers-2019-sentence-bert} to embed the Sentence T and Paraphrase R  and compute cosine similarity between them. We form a valid candidate set by retaining only those paraphrases that has threshold above 0.75 and hence making sure the meaning is preserved. From now on, candidate set refers to valid candidate set. Few examples of paraphrases can be seen in the table \ref{tab:freq}.

\section{Experiment}
\subsection{Dataset}
For training and evaluation purpose, we use Quora Question Pair(QQP) dataset\footnote{https://www.kaggle.com/c/quora-question-pairs}. QQP dataset has ~400k labelled examples out of which ~140k examples are actual paraphrases, and ~300k unlabeled sentence pairs. Each entry in QQP dataset has ($question1$, $question2$, $is\_duplicate$) fields where $question1$ and $question2$ are sentences and $is\_duplicate$ indicates whether or not they are paraphrases. Out of 140k actual paraphrase sentence pairs, we take random 30k sentence pairs for our test set as done in \cite{Liu2019UnsupervisedPB} and \cite{Miao2018CGMHCS} for the sake of easier comparison. We take all the unique question1 from remaining 370k labelled and 300k unlabeled sample giving total of 523010 unique sentences. We take 463010 sentences for training and 60000 sentences for validating the sentence reconstruction task. 

To demonstrate the performance enhancement of using paraphrasing for data augmentation, we use Stanford Sentiment Treebank (SST-2) dataset \cite{socher-etal-2013-recursive} as done in \cite{roy-grangier-2019-unsupervised} which has 6920 train examples and 1821 test examples.

\subsection{Training}
We initialize our model with GPT-2 medium checkpoint which has 345M parameters. The sentences are corrupted by removing the stop words. The model is finetuned on sentence reconstruction task the same way GPT-2 was trained in \cite{radford2019language}. validation set's perplexity is used to do early stopping. The model is fine-tuned on one Nvidia V100 DGX machine for 2-5 epochs which takes ~28 mins per epoch. 

\subsection{Evaluation Metric}
We use different evaluation metrics to measure quality, correctness, diversity and usefulness of the generated paraphrases. 

\begin{itemize}
  \item Quality : We use ROUGE-L \cite{lin-2004-rouge} and METEOR \cite{lavie-agarwal-2007-meteor} for measuring the quality of generated paraphrase. while BLEU-n calculates n gram overlap, ROUGE-L measures longest matching sequence, and it does not require consecutive matches but in-sequence matches that reflect sentence level word order. METEOR compares the texts using not just word matching, but stemming and synonym matching, which is more desirable to gauge in paraphrasing quality.
  
  \item Diversity : To measure the diversity of candidate paraphrases, we use self-BLEU from \cite{zhu-etal-2018-lingke}. Given set of paraphrase candidates for an input sentence, every candidate is picked once and others are used as references to compute BLEU score. And the average of all the BLEU scores computed for this candidate set is self-BLEU. A lower self-BLEU score indicates better diversity. 

  \item Usefulness : Using paraphrases for data augmentation for a downstream task can answer questions such as how good/diverse are paraphrases and do they provide any signal that was not provided by the original sentence and hence improving the downstream task's performance. We report \% increment in performance of the downstream task.
  
  \item Correctness : Though we filter the paraphrase candidates using high threshold on cosine similarity, it doesn't promise 100\% correctness. To measure what percent of the generated paraphrases are semantically similar to input sentence, we make human evaluation of the model output.
\end{itemize}

\begin{table*}[t!]
\centering
\caption{Paraphrasing Results - Compared with Supervised models}
\label{tab:results}
\begin{tabular}{ccc}
%\multirow{\textbf{Dataset}} & \multirow{\textbf{Model}} 
\hline
\textbf{Model} & $\uparrow$METEOR & $\downarrow$self-BLEU4\\
\hline \hline
 D-PAGE  & 28.54 & 85.41 \\
 PG-BS &  28.88 & 61.89\\
 qian-etal-2019\cite{qian-etal-2019-exploring}& 29.28 &  40.55\\
 \textbf{Ours(best candidate)}  & 50.48  & N/A\\
 \textbf{Ours(taking top 3 candidates)}  & 49.45 & 30.27\\
\hline 
\hline
  \multicolumn{2}{c}{\textbf{Our Model Human Evaluation: (Average  Accuracy)}  =} & \textbf{75.5\%}\\
 \hline\hline
\end{tabular}
\end{table*}

\begin{table}[t]
\centering
    \caption{Paraphrasing Results - Compared with unsupervised and unsupervised  models}
    \label{tab:unsup_results}
    \begin{tabular}{cccc}
        \hline
        & \textbf{Model}  & $\uparrow$ROUGE-1 & $\uparrow$ROUGE-2 \\
        \hline \hline
        & ResidualLSTM   & 59.22  & 32.40 \\
        Supervised & VAE-SVG-eq  & 59.98  & 33.30 \\
        & Pointer-generator   & 61.96  & 36.07 \\
        & Transformer   & 60.25  & 33.45 \\
        & Transformer+copy   & 63.34  & 37.31 \\
        & DNPG  & 63.73  & 37.75 \\
        \hline\hline
        & VAE & 44.55  & 22.64 \\
         & CGMH  & 48.73 & 26.12 \\
        Unsupervised &UPSA & 56.51 & 30.69\\
        & \textbf{Ours(best candidate)} &\textbf{60.33}  & \textbf{34.59} \\
         & \textbf{Ours(top 3 candidates)} &\textbf{59.43}  & \textbf{33.61} \\
        \hline

    \end{tabular}
\end{table}

\begin{table}[t]
\centering
    \caption{Data Augmentation for Classification }
    \label{tab:augment}
    \begin{tabular}{ccccccc}
        \hline\hline
        & &\multicolumn{2}{c}{SST-2}
        \\\cline{3-4}
        Model & Metric &Acc.& F1     \\
        \hline\hline
        & Roy and Grangier(2019)\cite{roy-grangier-2019-unsupervised}: Baseline & 81.93 & 83.15\\
        NB-SVM (trigram) & Roy and Grangier(2019)\cite{roy-grangier-2019-unsupervised} Enhanced & 82.12 & 83.23  \\
        & \% increment & 0.23\% & 0.1\%\\
        \hline

        & Ours  Baseline & 77.14 & 77.56  \\
        NB-SVM (trigram) & Ours Enhanced  & 79.39 & 79.71 \\
        & \textbf{Ours \% increment} & \textbf{2.92\%} & \textbf{2.77\%} \\
        \hline

        & Ours  Baseline & 0.62 & 0.62 \\
        TFIDF+RF & Ours  Enhanced & 0.68 & 0.67 \\
        & \textbf{Ours \% increment} & \textbf{9.68\%} & \textbf{8.06\%}\\
        \hline\hline
    \end{tabular}
\end{table}

\subsection{Results}
\subsubsection{Quality and Diversity}
We compute METEOR between generated paraphrase R and $question2$ from QQP test set for all 30k examples. We conduct multiple experiments with respect to final paraphrase selection from candidate paraphrases. In one of the experiments, we only take the best one out of all the candidates, where as in the second one we compute values for each candidate against $question2$ and take the average of them. We have further calculate the self-BLEU among the paraphrase candidates. All these values can be seen in Table \ref{tab:results}. 

We compare our results with both supervised and un-supervised methods. The selective supervised methods we compare our method to in table \ref{tab:results}, are D-PAGE \cite{Xu2018DPAGEDP} and Pointer generator with Beam-search (PG-BS) \cite{See2017GetTT} and \cite{qian-etal-2019-exploring}. Values seen in the table \ref{tab:results} are reported as per \cite{qian-etal-2019-exploring} for all these methods. The unsupervised methods that we compare our results with are VAE \cite{Bowman2016GeneratingSF}, CGMH \cite{Miao2018CGMHCS} and UPSA \cite{Liu2019UnsupervisedPB}. The values for these papers seen in table \ref{tab:unsup_results} are taken from \cite{Liu2019UnsupervisedPB}.

While we don't have access to exact test set used in \cite{Liu2019UnsupervisedPB}, we follow the exact guideline from \cite{Liu2019UnsupervisedPB}, \cite{Miao2018CGMHCS} and \cite{qian-etal-2019-exploring} where they mention taking random 30k sample out of ~140k paraphrases form QQP dataset.

\subsubsection{Usefulness}
To demonstrate the usefulness of paraphrases on a downstream task, we trained NB-SVM and Random Forest Classifier on SST-2 and TREC dataset. We have reported the percentage change in accuracy before and after using paraphrases for data augmentation. Table \ref{tab:augment} shows that the improvement in performance is significant when compared to \cite{roy-grangier-2019-unsupervised}. Our baseline results doesn't match with \cite{roy-grangier-2019-unsupervised} but absolute numbers shouldn't matter as we keep all the settings same before and after data augmentation for measuring the \% increment except for the training data.

\subsubsection{Human Evaluation for Correctness: Higher[/Lower] BLEU does "not" mean better[/worse] Paraphrase}
BLEU score measures how similar is the candidate text is to the reference texts by looking at the overlapping n-grams. If an example $x_i$ has lower BLEU when compared against the paraphrase given in evaluation set, it doesn't mean that $x_i$ is not the correct paraphrase, but it means that $x_i$ has different set of tokens than the reference sentence from evaluation set. While self-BLUE captures diversity, we use human evaluation to measure correctness.

We form two human evaluation sets containing 100 examples each from QQP dataset having model generated paraphrase sentence R and ground truth paraphrase sentence($question2$), we asked 2 humans who are fluent in English to label whether or not R and $question2$ are paraphrases. Each of the annotators were asked to label it as 0 or 1, where Label=0 means not a paraphrase and Label=1 means they are paraphrases. We compute average accuracy and saw that 75.5\% of the time, model generated paraphrase was correct.

\subsubsection{Takeaway}
Table \ref{tab:results} shows that our model has significantly higher METEOR score than supervised models, highlighting the higher quality of paraphrases. Low self-BLEU among candidates generated by our model indicate that the candidate sentences are more diverse than the candidate paraphrases generated by other supervised methods. table \ref{tab:unsup_results} compares quality of paraphrase generated by our model with more model performances on Quora dataset with ROUGE metric,and our model does better in comparison to all unsupervised methods here as well, with being very much closely comparable to the supervised methods. Human evaluation of the paraphrases by our model shows them to be correct 75.5\% of the time. We further show that the generated paraphrases are not redundant but useful by showing improvement in performance on two tasks using two models when paraphrases are used for data augmentation on a downstream task.
Finally, in table \ref{tab:augment} we report that augmenting the training set of SST-2 dataset with generated paraphrases from our model, sees much higher gain as compared to experiments reported in \cite{roy-grangier-2019-unsupervised}.

\section{Conclusion and Future Work}
In this paper, we have introduced an unsupervised paraphrasing model that generates good quality, diverse and helpful paraphrases. We demonstrate the usefulness of paraphrases by using them to augment the training data of downstream tasks. Since the model doesn't need labelled data, it can be trained as sentence reconstruction task on independent sentences pertaining to any domain without suffering from domain shift problem faced by most of the supervised models.

The sentence corruption technique has a lot of scope for further exploration and improvement, and as future work, we'd like to explore more complex and sophisticated way of corrupting the input sentence, and it's effect on the robustness in terms of reconstruction and generated paraphrases. We have seen that number of epochs for training the model is a trade-off between correctness and diversity, and we'd like to investigate more on that too in subsequent experiments.

\bibliographystyle{unsrt}  
% \bibliography{references}  

\begin{thebibliography}{10}

\bibitem{radford2019language}
Alec Radford, Jeff Wu, Rewon Child, David Luan, Dario Amodei, and Ilya
  Sutskever.
\newblock Language models are unsupervised multitask learners.
\newblock 2019.

\bibitem{koehn-knowles-2017-six}
Philipp Koehn and Rebecca Knowles.
\newblock Six challenges for neural machine translation.
\newblock In {\em Proceedings of the First Workshop on Neural Machine
  Translation}, pages 28--39, Vancouver, August 2017. Association for
  Computational Linguistics.

\bibitem{barzilay-lee-2003-learning}
Regina Barzilay and Lillian Lee.
\newblock Learning to paraphrase: An unsupervised approach using
  multiple-sequence alignment.
\newblock In {\em Proceedings of the 2003 Human Language Technology Conference
  of the North {A}merican Chapter of the Association for Computational
  Linguistics}, pages 16--23, 2003.

\bibitem{quirk-etal-2004-monolingual}
Chris Quirk, Chris Brockett, and William Dolan.
\newblock Monolingual machine translation for paraphrase generation.
\newblock In {\em Proceedings of the 2004 Conference on Empirical Methods in
  Natural Language Processing}, pages 142--149, Barcelona, Spain, July 2004.
  Association for Computational Linguistics.

\bibitem{duboue-chu-carroll-2006-answering}
Pablo Duboue and Jennifer Chu-Carroll.
\newblock Answering the question you wish they had asked: The impact of
  paraphrasing for question answering.
\newblock In {\em Proceedings of the Human Language Technology Conference of
  the {NAACL}, Companion Volume: Short Papers}, pages 33--36, New York City,
  USA, June 2006. Association for Computational Linguistics.

\bibitem{10.5555/3298023.3298027}
Ziqiang Cao, Chuwei Luo, Wenjie Li, and Sujian Li.
\newblock Joint copying and restricted generation for paraphrase.
\newblock In {\em Proceedings of the Thirty-First AAAI Conference on Artificial
  Intelligence}, AAAI’17, page 3152–3158. AAAI Press, 2017.

\bibitem{prakash-etal-2016-neural}
Aaditya Prakash, Sadid~A. Hasan, Kathy Lee, Vivek Datla, Ashequl Qadir, Joey
  Liu, and Oladimeji Farri.
\newblock Neural paraphrase generation with stacked residual {LSTM} networks.
\newblock In {\em Proceedings of {COLING} 2016, the 26th International
  Conference on Computational Linguistics: Technical Papers}, pages 2923--2934,
  Osaka, Japan, December 2016. The COLING 2016 Organizing Committee.

\bibitem{su-yan-2017-cross}
Yu~Su and Xifeng Yan.
\newblock Cross-domain semantic parsing via paraphrasing.
\newblock In {\em Proceedings of the 2017 Conference on Empirical Methods in
  Natural Language Processing}, pages 1235--1246, Copenhagen, Denmark,
  September 2017. Association for Computational Linguistics.

\bibitem{Buck2017AskTR}
Christian Buck, Jannis Bulian, Massimiliano Ciaramita, Andrea Gesmundo, Neil
  Houlsby, Wojciech Gajewski, and Wei Wang.
\newblock Ask the right questions: Active question reformulation with
  reinforcement learning.
\newblock {\em ArXiv}, abs/1705.07830, 2017.

\bibitem{Dong2017LearningTP}
Li~Dong, Jonathan Mallinson, Siva Reddy, and Mirella Lapata.
\newblock Learning to paraphrase for question answering.
\newblock {\em ArXiv}, abs/1708.06022, 2017.

\bibitem{Fu2019ParaphraseGW}
Yao Fu, Yansong Feng, and John~P. Cunningham.
\newblock Paraphrase generation with latent bag of words.
\newblock {\em ArXiv}, abs/2001.01941, 2019.

\bibitem{Gupta2018ADG}
Ankush Gupta, Arvind Agarwal, Prawaan Singh, and Piyush Rai.
\newblock A deep generative framework for paraphrase generation.
\newblock {\em ArXiv}, abs/1709.05074, 2018.

\bibitem{Li2018ParaphraseGW}
Zichao Li, Xin Jiang, Lifeng Shang, and Huang Li.
\newblock Paraphrase generation with deep reinforcement learning.
\newblock In {\em EMNLP}, 2018.

\bibitem{nomoto-2019-generating}
Tadashi Nomoto.
\newblock Generating paraphrases with lean vocabulary.
\newblock In {\em Proceedings of the 12th International Conference on Natural
  Language Generation}, pages 438--442, Tokyo, Japan, October{--}November 2019.
  Association for Computational Linguistics.

\bibitem{qian-etal-2019-exploring}
Lihua Qian, Lin Qiu, Weinan Zhang, Xin Jiang, and Yong Yu.
\newblock Exploring diverse expressions for paraphrase generation.
\newblock In {\em Proceedings of the 2019 Conference on Empirical Methods in
  Natural Language Processing and the 9th International Joint Conference on
  Natural Language Processing (EMNLP-IJCNLP)}, pages 3173--3182, Hong Kong,
  China, November 2019. Association for Computational Linguistics.

\bibitem{Iyyer2018AdversarialEG}
Mohit Iyyer, John Wieting, Kevin Gimpel, and Luke Zettlemoyer.
\newblock Adversarial example generation with syntactically controlled
  paraphrase networks.
\newblock In {\em NAACL-HLT}, 2018.

\bibitem{Bowman2016GeneratingSF}
Samuel~R. Bowman, Luke Vilnis, Oriol Vinyals, Andrew~M. Dai, Rafal
  J{\'o}zefowicz, and Samy Bengio.
\newblock Generating sentences from a continuous space.
\newblock In {\em CoNLL}, 2016.

\bibitem{Miao2018CGMHCS}
Ning Miao, Hao Zhou, Lili Mou, Rui Yan, and Lei Li.
\newblock Cgmh: Constrained sentence generation by metropolis-hastings
  sampling.
\newblock In {\em AAAI}, 2018.

\bibitem{roy-grangier-2019-unsupervised}
Aurko Roy and David Grangier.
\newblock Unsupervised paraphrasing without translation.
\newblock In {\em Proceedings of the 57th Annual Meeting of the Association for
  Computational Linguistics}, pages 6033--6039, Florence, Italy, July 2019.
  Association for Computational Linguistics.

\bibitem{Oord2017NeuralDR}
A{\"a}ron van~den Oord, Oriol Vinyals, and Koray Kavukcuoglu.
\newblock Neural discrete representation learning.
\newblock In {\em NIPS}, 2017.

\bibitem{Liu2019UnsupervisedPB}
Xianggen Liu, Lili Mou, Fandong Meng, Hao Zhou, Jie Zhou, and Sen Song.
\newblock Unsupervised paraphrasing by simulated annealing.
\newblock {\em ArXiv}, abs/1909.03588, 2019.

\bibitem{Witteveen2019ParaphrasingWL}
Sam Witteveen and Martin Andrews.
\newblock Paraphrasing with large language models.
\newblock In {\em NGT@EMNLP-IJCNLP}, 2019.

\bibitem{Wolf2019HuggingFacesTS}
Thomas Wolf, Lysandre Debut, Victor Sanh, Julien Chaumond, Clement Delangue,
  Anthony Moi, Pierric Cistac, Tim Rault, R'emi Louf, Morgan Funtowicz, and
  Jamie Brew.
\newblock Huggingface's transformers: State-of-the-art natural language
  processing.
\newblock {\em ArXiv}, abs/1910.03771, 2019.

\bibitem{fan-etal-2018-hierarchical}
Angela Fan, Mike Lewis, and Yann Dauphin.
\newblock Hierarchical neural story generation.
\newblock In {\em Proceedings of the 56th Annual Meeting of the Association for
  Computational Linguistics (Volume 1: Long Papers)}, pages 889--898,
  Melbourne, Australia, July 2018. Association for Computational Linguistics.

\bibitem{reimers-2019-sentence-bert}
Nils Reimers and Iryna Gurevych.
\newblock Sentence-bert: Sentence embeddings using siamese bert-networks.
\newblock In {\em Proceedings of the 2019 Conference on Empirical Methods in
  Natural Language Processing}. Association for Computational Linguistics, 11
  2019.

\bibitem{socher-etal-2013-recursive}
Richard Socher, Alex Perelygin, Jean Wu, Jason Chuang, Christopher~D. Manning,
  Andrew Ng, and Christopher Potts.
\newblock Recursive deep models for semantic compositionality over a sentiment
  treebank.
\newblock In {\em Proceedings of the 2013 Conference on Empirical Methods in
  Natural Language Processing}, pages 1631--1642, Seattle, Washington, USA,
  October 2013. Association for Computational Linguistics.

\bibitem{lin-2004-rouge}
Chin-Yew Lin.
\newblock {ROUGE}: A package for automatic evaluation of summaries.
\newblock In {\em Text Summarization Branches Out}, pages 74--81, Barcelona,
  Spain, July 2004. Association for Computational Linguistics.

\bibitem{lavie-agarwal-2007-meteor}
Alon Lavie and Abhaya Agarwal.
\newblock {METEOR}: An automatic metric for {MT} evaluation with high levels of
  correlation with human judgments.
\newblock In {\em Proceedings of the Second Workshop on Statistical Machine
  Translation}, pages 228--231, Prague, Czech Republic, June 2007. Association
  for Computational Linguistics.

\bibitem{zhu-etal-2018-lingke}
Pengfei Zhu, Zhuosheng Zhang, Jiangtong Li, Yafang Huang, and Hai Zhao.
\newblock {L}ingke: a fine-grained multi-turn chatbot for customer service.
\newblock In {\em Proceedings of the 27th International Conference on
  Computational Linguistics: System Demonstrations}, pages 108--112, Santa Fe,
  New Mexico, August 2018. Association for Computational Linguistics.

\bibitem{Xu2018DPAGEDP}
Qiongkai Xu, Juyan Zhang, Lizhen Qu, Lexing Xie, and Richard Nock.
\newblock D-page: Diverse paraphrase generation.
\newblock {\em ArXiv}, abs/1808.04364, 2018.

\bibitem{See2017GetTT}
Abigail See, Peter~J. Liu, and Christopher~D. Manning.
\newblock Get to the point: Summarization with pointer-generator networks.
\newblock {\em ArXiv}, abs/1704.04368, 2017.

\end{thebibliography}
%%% Remove comment to use the external .bib file (using bibtex).
%%% and comment out the ``thebibliography'' section.

\end{document}